\documentclass[conference]{IEEEtran}
\IEEEoverridecommandlockouts
\usepackage{cite}
\usepackage{multirow}
\usepackage{hyperref}
\usepackage{amsmath,amssymb,amsfonts}
\usepackage{algorithmic}
\usepackage{graphicx}
\usepackage{textcomp}
\usepackage{xcolor}
\def\BibTeX{{\rm B\kern-.05em{\sc i\kern-.025em b}\kern-.08em
    T\kern-.1667em\lower.7ex\hbox{E}\kern-.125emX}}
\begin{document}

\title{Landmark-guided Diffusion Model for High-fidelity and Temporally Coherent Talking Head Generation}


\author{
Jintao Tan$^{1,4}$*\thanks{*Equal contribution.} , 
Xize Cheng$^2$*,
Lingyu Xiong$^{1,4}$* \\
Lei Zhu$^3$ ,
Xiandong Li$^4$ ,
Xianjia Wu$^4$ ,
Kai Gong$^4$ ,
Minglei Li$^4$,
Yi Cai$^1$$^\dag\thanks{\dag Corresponding author.}$\\
$^1$South China University of Technology\quad $^2$Zhejiang University \\
$^3$Peking University\quad $^4$Huawei Cloud
}
\maketitle

\begin{abstract}
Audio-driven talking head generation is a significant and challenging task applicable to various fields such as virtual avatars, film production, and online conferences. However, the existing GAN-based models emphasize generating well-synchronized lip shapes but overlook the visual quality of generated frames, while diffusion-based models prioritize generating high-quality frames but neglect lip shape matching, resulting in jittery mouth movements. To address the aforementioned problems, we introduce a two-stage diffusion-based model. The first stage involves generating synchronized facial landmarks based on the given speech. In the second stage, these generated landmarks serve as a condition in the denoising process, aiming to optimize mouth jitter issues and generate high-fidelity, well-synchronized, and temporally coherent talking head videos. Extensive experiments demonstrate that our model yields the best performance.
\end{abstract}


\section{Introduction}
Talking Head Generation (THG) stands as a significant and promising research domain within AIGC. Its objective is to synthesize a talking head video\cite{b1,b2,b3,b4,b5} guided by provided speech, finding wide application across various fields such as virtual avatars, film production, and online conferences \cite{b6,b7,b8}. In THG, researchers primarily emphasize three key aspects: lip-speech synchronization, visual quality, and temporal coherence. 
The primary objective of THG is to create talking head videos where lip movements align precisely with speech, hence making lip-speech synchronization a pivotal metric. The human sensitivity to video clarity underscores the importance of visual quality in the generated frames, significantly impacting the overall video quality. Beyond clarity, the coherence between frames plays a crucial role in establishing the authenticity and fluidity of the video. Hence, attention to the video's temporal coherence is equally essential.

Recent years have seen the emergence of several commendable works in this research field, primarily concentrating on optimizing the aforementioned three aspects. Within these endeavors, certain GAN-based methodologies \cite{b1,b2,b9} demonstrate the capability to generate talking head videos with highly accurate lip-speech synchronization. Nevertheless, as these methods generate facial images separately and subsequently integrate them back into the original frames, artifacts or false details along the edges of the integrated areas may arise, potentially affecting the authenticity of the resulting video. Moreover, the inherent characteristics of GANs contribute to unstable training, heightened sensitivity to hyperparameter selection, and a susceptibility to issues like mode collapse. Conversely, diffusion-based methods \cite{b3,b10,b11} excel in producing artifact-free frames with high visual quality. However, owing to the diverse range of images generated by the diffusion model, there might be a trade-off, potentially leading to compromised temporal coherence within the resulting video.

\begin{figure*}
  \centering
  \setlength{\abovecaptionskip}{0.5em}
  \includegraphics[width=1\linewidth]{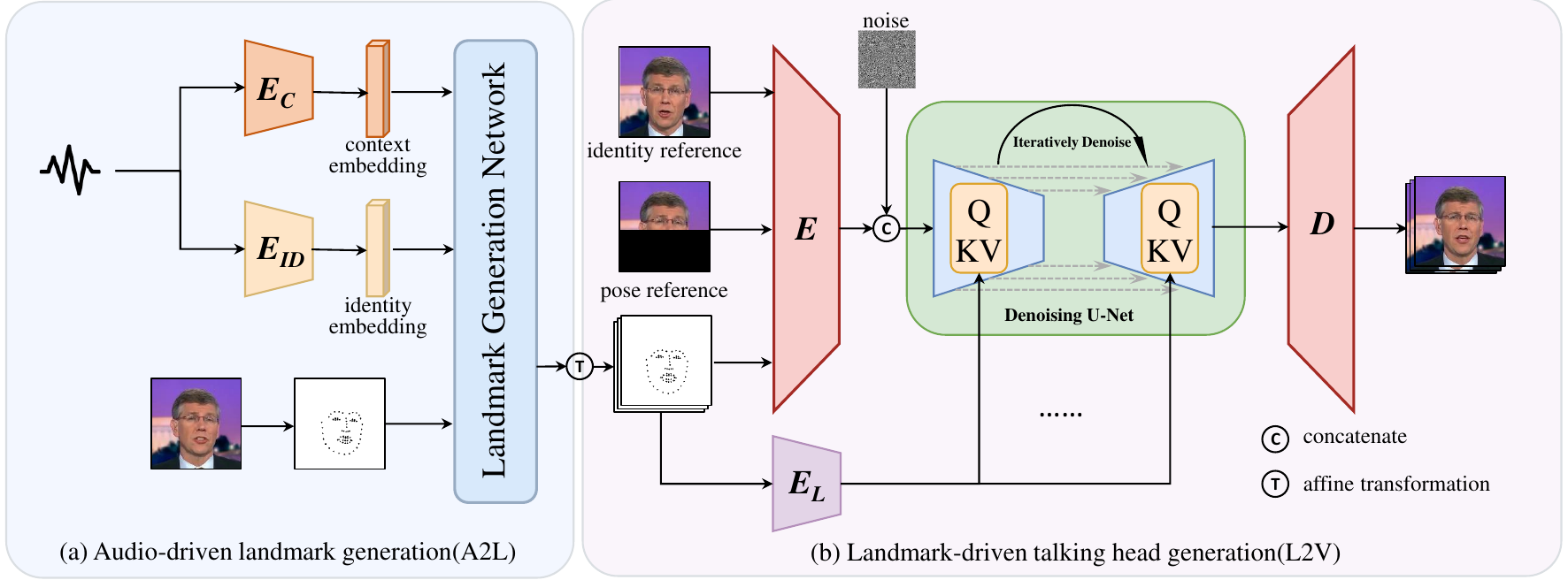}
  \caption{Overview of our proposed model for talking head generation, which consists of two sub-modules. (a)Audio-driven landmark generation (A2L) module takes as input the given speech and the original facial image to generate the landmark movement sequence. (b)The generated landmarks will serve as a condition for the denoising process in landmark-driven talking head generation (L2V) module.}
  \label{fig:talking_face}
\end{figure*}

To leverage the outstanding image generation capabilities of diffusion models \cite{b12,b13,b14}  and ensure the temporal consistency of the resulting videos, we introduce a novel two-stage diffusion-based model. 
Figure.\ref{fig:talking_face} illustrates our approach, where we segment the talking head synchronization task into two stages, bridged by facial landmarks serving as the intermediary representation. In the first stage, we generate facial landmarks guided by the given speech. Then, in the second stage, these generated landmarks serve as guiding conditions for the denoising process. Using facial landmarks instead of relying solely on the given speech as a condition offers more robust guidance for the denoising process. This approach ensures enhanced temporal consistency in the resulting videos. 

In summary, our contributions are as follows:
\begin{itemize}
\setlength{\itemsep}{0pt}
\item We propose the first landmark-guided diffusion model for talking head generation.
\item We devise a two-stage framework that notably enhances the temporal coherence of talking head videos while preserving the visual quality of frames.
\item Extensive experiments demonstrate that our approach yields the best performance. Please refer to the supplementary materials for related results. 
\end{itemize}

\section{METHOD}

\subsection{Overview}
To harness the outstanding attributes of Latent Diffusion Models in producing high-quality images swiftly, and to address mouth jitter in the generation of talking faces, we introduce a novel two-stage diffusion-based model. Figure.\ref{fig:talking_face} provides an overview of the proposed model. The initial phase involves the Landmark Generation Network, which utilizes the provided speech and the original facial landmark to produce the landmark movement sequence. This sequence acts as a condition for the subsequent denoising process. In the second stage, the denoising process is guided by utilizing the identity reference image, the pose reference image, and the generated landmarks as constraints. This process aims to generate a high-fidelity, synchronized talking head video. Subsequently, we will provide detailed elaboration on the two aforementioned sub-modules in Sec.\ref{sub:audio-driven} and Sec.\ref{sub:landmark-driven}. 

\subsection{Audio-driven landmark generation}
\label{sub:audio-driven}

The first sub-module is the audio-driven landmark generation, called A2L. Given an audio clip and a facial image,  A2L will generate a sequence of landmark movements that align with the audio. For the landmark part, we employ an existing network \cite{b15} to extract the 2D facial landmark, denoted as \textit{l} $\in$ $\mathbb{R}^{68\times2}$. Speech contains two types of information, i.e., identity information and context information. Therefore, two different encoders are employed to extract identity embedding and context embedding. We utilize AutoVC network \cite{b16}, denoted as $\textit{E}_c$, to extract the speaker-agnostic content embedding \textit{$a_c$} $\in$ $\mathbb{R}^{T\times {D}_{1}}$, where \textit{T} is the total number of input audio frames, and $\textit{D}_{1}$ is the content dimension. We utilize a speaker verification model \cite{b17}, denoted as $\textit{E}_{id}$, to extract the speaker identity embedding \textit{$a_{id}$} $\in$ $\mathbb{R}^{{D}_{2}}$, where $\textit{D}_{2}$ is the identity dimension.

The Landmark Generation Network comprises two LSTM modules: one serving as the context network and the other as the identity network. The context network takes as input the context embedding of the audio and the original landmark \textit{l} to generate a speaker-independent sequence of landmark movements, which can be formulated as follows:
\begin{equation}
    \overline{l}_{i} = Context(a_{c,t\to t_{i}},l )
    \label{eq:eq1}
\end{equation}
Where $\overline{l}_{i}$ is the $\textit{i}$-th speaker-independent intermediate landmark, solely reflecting the content of the speech. $a_{c,t\to t_{i}}$ represents the context embedding from frame \textit{t} to frame $\textit{t}_{i}$, aligned with the $\textit{i}$-th landmark.

The intermediate landmarks contain only motion information relevant to the speech content but lack personal style. To further refine the intermediate landmarks and generate an accurate sequence of landmark movements, the identity network is employed. After the refinement process, the generated landmarks will be closer to the actual facial landmarks. The refinement process can be formulated as follows:
\begin{equation}
    \tilde{l}_{i} = Identity(a_{c,t\to t_{i}},a_{id},\overline{l}_{i} )
    \label{eq:eq2}
\end{equation}
Where $a_{id}$ is the identity embedding extracted by the audio encoder $\textit{E}_{id}$. $\tilde{l}_{i}$ is the finally generated landmarks with personal style that match the given audio. which will serve as a condition to guide the denoising process in the second sub-module. 

\textbf{Loss fuction}. We only train the Landmark Generation Network, while the two audio encoders are frozen. We optimize the Landmark Generation Network by minimizing the mean of the square difference between the predicted landmarks and the ground truth. The loss function can be formulated as follows:
\begin{equation}
    \mathcal{L} = \sum_{i=1}^{K} \sum_{j=1}^{N}\left \| \tilde{l}_{i,j}-\tilde{l}_{i,j}^{gt}  \right \|_2^2  
    \label{eq:eq3}
\end{equation}
Where $\tilde{l}_{i,j}$ is the $\textit{j}$-th point of the $\textit{i}$-th predicted landmark and $\tilde{l}_{i,j}^{gt}$ is the $\textit{j}$-th point of the $\textit{i}$-th ground truth landmark. \textit{N} is equal to 68, representing the number of all points in one facial landmark. \textit{K} is the number of predicted landmarks.

\subsection{Landmark-driven talking head generation}
\label{sub:landmark-driven}

\textbf{Diffusion Models}. Diffusion Models (DMs) \cite{b18} work by destroying training data through the successive addition of Gaussian noise and then learning to recover the data by reversing this noising process, which corresponds to learning the reverse process of a fixed Markov Chain of length \textit{T}. They can be considered as an equally weighted sequence of denoising autoencoders $\epsilon _{\theta } (x_{t},t)$. The optimization process can be defined as follows:
\begin{equation}
    \textit{L}_{DM} = \mathbb{E}_{x,\epsilon \sim \mathcal{N}(0,1),t}\left [\left \| \epsilon -\epsilon _{\theta}(x_t,t)\right \|^2_2   \right ]
    \label{eq:DM}
\end{equation}
where \textit{t} is sampled from $\left \{ 1,\dots ,T \right \} $ and $x_{t}$ is a noisy version of the input $x$.

\noindent\textbf{Latent Diffusion Models}. 
The denoising process of the diffusion models is operated in high-dimensional pixel space,  resulting in high costs for both training and inference. To generate high-quality images while reducing training costs, Latent Diffusion Models(LDMs) emerged based on DMs. To reduce the computational complexity of the model, LDMs introduce the perceptual image compression method. It consists of an autoencoder, which includes an encoder \textit{E} and a decoder \textit{D}. Given an input image $x$, the encoder \textit{E} encodes it to a latent representation $z = E(x)$, and then the decoder \textit{D} reconstructs the image from latent: $\tilde{x}=D(z)$. The denoising process of LDMs is operated in this low-dimensional latent space in which high-frequency imperceptible details are abstracted away, resulting in an effective and efficient way for high-quality image synthesis. The objective of LDMs can be formulated as follows:
\begin{equation}
    \textit{L}_{LDM} = \mathbb{E}_{E(x),\epsilon \sim \mathcal{N}(0,1),t}\left [\left \| \epsilon -\epsilon _{\theta}(z_t,t)\right \|^2_2   \right ]
    \label{eq:eq5}
\end{equation}
The denoising network $\epsilon_\theta$ is realized as a time-conditional UNet \cite{b19}. $z_t$ is obtained through the forward diffusion process from $z = E(x)$.

\noindent\textbf{Conditioning Mechanisms}. The Landmark-driven talking head generation (L2V), as shown in Figure.\ref{fig:talking_face}, is based on LDMs. 
We use a well-trained autoencoder model \cite{b20} as the encoder \textit{E} and decoder \textit{D} in L2V. 
There are three types of conditions to guide the denoising process: the pose reference image $x_p\in \mathbb{R}^{H\times W\times 3}$, the identity reference image $x_{id}\in \mathbb{R}^{H\times W\times 3}$, and the landmark $l$. During training, the landmark $l$ is extracted from the ground truth image $x$. During inference, it's derived from the landmark predicted by A2L after an affine transformation. Similar to the setup in Wav2Lip \cite{b2}, $x_{id}$ is used to provide identity information, while $x_{p}$ is used to offer accurate pose information, enabling the model to focus on synthesizing lip shapes. Before serving as conditions, $x_{id}$ and $x_{p}$ are encoded into the latent space by encoder \textit{E}: $z_{id} = E(x_{id})\in \mathbb{R}^{h\times w\times 3}$, $z_{p} = E(x_{p})\in \mathbb{R}^{h\times w\times 3}$, where $H/h=W/w=f$, $H,W$ are the height and width of the original image and $f$ is the downsampling factor. The objective of L2V can be formulated as follows:
\begin{equation}
    \textit{L} = \mathbb{E}_{E(x),\epsilon \sim \mathcal{N}(0,1),t}\left [\left \| \epsilon -\epsilon _{\theta}(z_t,t,z_p,z_{id},l)\right \|^2_2   \right ]
    \label{eq:eq6}
\end{equation}
More specifically, the input for the landmark condition comprises two categories: the visual input and the coordinate input. The subsequent ablation study can demonstrate the effectiveness of these two types of inputs. For the visual input, we convert the
landmark $l$ into the image $x_l\in \mathbb{R}^{H\times W\times 3}$ and then encode it into the latent space: $z_{l} = E(x_{l})\in \mathbb{R}^{h\times w\times 3}$. $z_{l}$, $z_{p}$, $z_{id}$, and the noise map $z_{T}$ are concatenated together to serve as the query for the intermediate cross-attention layers of the denoising network $\epsilon_\theta$. For the coordinate input, we use the landmark encoder $E_L$ to extract the coordinate embedding $C_l=E_L(l)\in\mathbb{R}^{D_l}$, which serves as the key and value in denoising network $\epsilon_\theta$.

\renewcommand{\dblfloatpagefraction}{.9}
\begin{figure*}
  \centering
  \setlength{\abovecaptionskip}{0.5em}
  \includegraphics[width=0.9\linewidth]{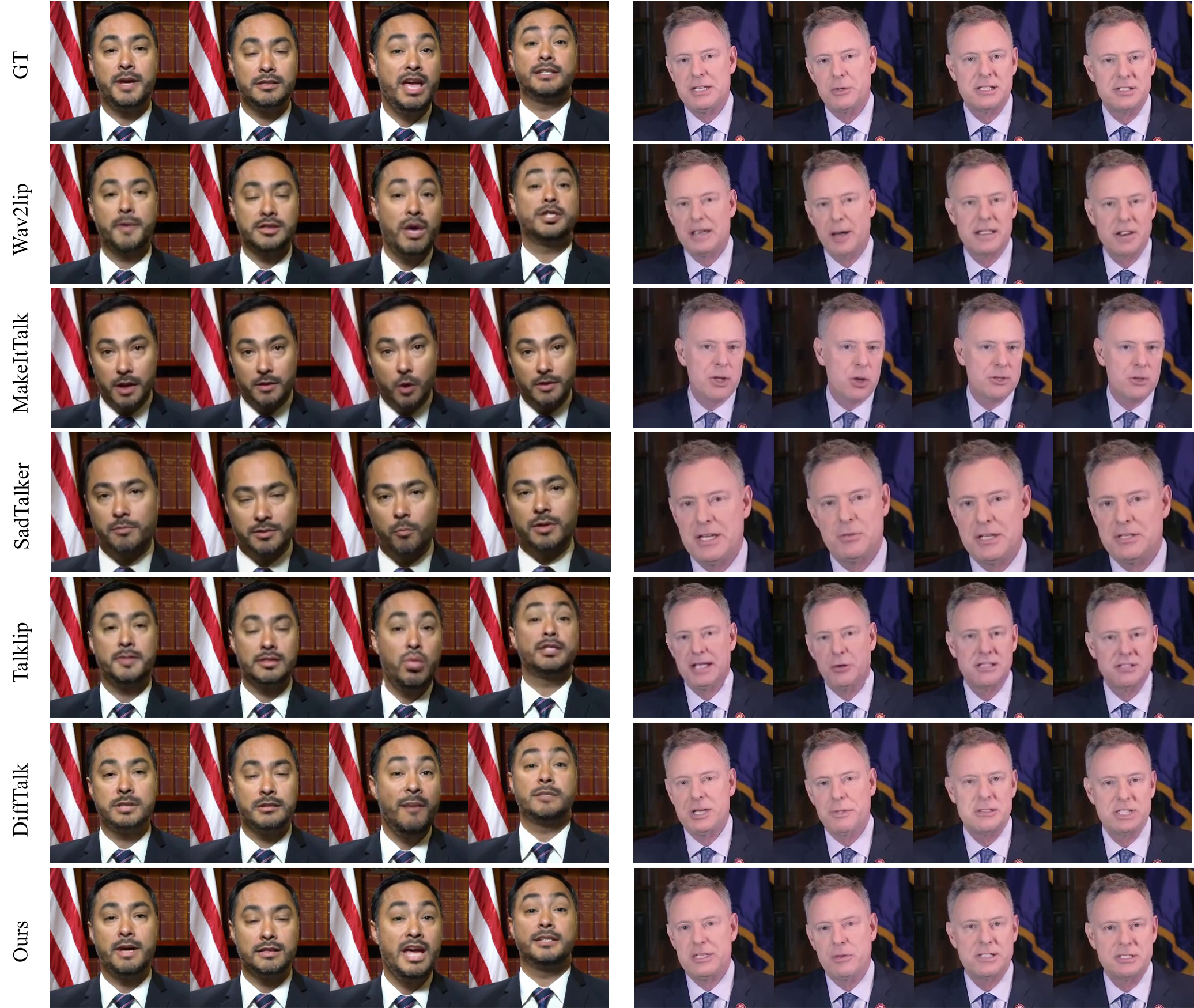}
  \caption{Visual comparisons between our proposed method and several state-of-the-art methods Wav2lip \cite{b2}, MakeItTalk \cite{b8}, Talklip \cite{b1}, SadTalker \cite{b21} and Difftalk \cite{b3}.}
  \label{fig:result}
\end{figure*}

\section{EXPERIMENTS}
\subsection{Experimental Settings}
\label{sub: Implementation Details}
\noindent\textbf{Datasets}.We use the audio-visual dataset HDTF \cite{b22} to train these two sub-modules A2L and L2V. It is gathered from YouTube, encompassing over three hundred distinct individuals, with a resolution of 720P or 1080P. It contains about 17 hours of talking videos. We randomly select 50 videos for training, totaling approximately 4 hours in duration, while the remaining videos are reserved for testing purposes.

\noindent\textbf{Metric}.We evaluate the proposed model from three perspectives, including lip synchronization, visual quality, and temporal coherence. For lip synchronization, we use LSE-C ($\uparrow$) and LSE-D ($\downarrow$), which are proposed to quantify the lip-speech synchronization using a well-trained Lip-Sync expert. For visual quality, we employ PSNR ($\uparrow$), SSIM ($\uparrow$), LPIPS ($\downarrow$), and FID ($\downarrow$) to evaluate our model. 
For temporal coherence, tLP ($\downarrow$) \cite{b23} and Pixel-MSE ($\downarrow$) are employed. tLP employs the perceptual LPIPS metric to measure the visual similarity of two consecutive frames. Pixel-MSE is used to calculate the averaged mean squared pixel error between aligned consecutive frames. (“$\downarrow$” indicates that the lower the better, while “$\uparrow$” means that the higher the better.)

\noindent\textbf{Implementation Details}.
We use PyTorch to implement our framework. In A2L, the content dimension $\textit{D}_{1}$ and the identity dimension $\textit{D}_{2}$ are both set to 256. In L2V, the height and width of the input images are resized to 256. The downsampling factor \textit{f} is set as 4, so the latent space is 64$\times$64$\times$3. The A2L module takes about 8 hours to train on one NVIDIA A100 GPU with 40GB, while the L2V module takes about 20 hours on two NVIDIA A100 GPUs. In L2V, the diffusion model is trained using 1000 diffusion steps. During the inference stage, we use 200 steps of DDIM sampling \cite{b24}.

\makeatletter
\patchcmd{\@makecaption}
  {\scshape}
  {}
  {}
  {}
\makeatletter
\patchcmd{\@makecaption}
  {\\}
  {.\ }
  {}
  {}
\makeatother
\def\tablename{Table}

\begin{table*}[t]
\centering
\caption{Quantitative comparisons between our proposed method and some representative talking head generation methods on the HDTF dataset. The best performance is highlighted \textbf{in bold}. Our method performs extremely well in terms of visual quality of the generated frames. It obtains the best FID, PSNR, SSIM, and LPIPS values, and comparable performance in lip synchronization.} 
\label{tab:cap1}
\begin{tabular}{c|cc|cccc}
\hline

\multirow{2}{*}{Method} & \multicolumn{2}{c|}{lip Synchronization} & \multicolumn{4}{c}{visual quality} \\ \cline{2-7} 
           & LSE-C$\uparrow$ & LSE-D$\downarrow$  & FID$\downarrow$    & LPIPS$\downarrow$  & PSNR$\uparrow$   & SSIM$\uparrow$  \\ \hline
GT         & 7.305 & 7.569  & -      & -      & -      & -     \\
Wav2lip \cite{b2}    & \textbf{8.299} & \textbf{7.124}  & 20.680 & 0.1552 & 31.134 & 0.921 \\
MakeItTalk \cite{b8} & 4.304 & 10.222 & 34.507 & 0.4569 & 20.115 & 0.631 \\
SadTalker \cite{b21}  & 6.382 & 8.397  & 86.101 & 0.9726 & 14.115 & 0.338 \\
Talklip \cite{b1}    & 7.675 & 7.185  & 22.210 & 0.1800 & 31.601 & 0.922 \\
Difftalk \cite{b3}   & 4.601 & 9.337 & 23.497 & 0.1109 & 31.653 & 0.913 \\
ours                       & 6.636 & 8.015  & \textbf{18.384} & \textbf{0.0853} & \textbf{32.140} & \textbf{0.928} \\ \hline
\end{tabular}
\end{table*}

\subsection{Main Results}
\noindent\textbf{Qualitative Analysis}.
To validate the effectiveness of our proposed model, we compare it with several state-of-the-art methods, namely Wav2lip \cite{b2}, MakeItTalk \cite{b8}, Talklip \cite{b1}, SadTalker \cite{b12}, and Difftalk \cite{b3}. We randomly select some generated frames for visualization, which can be seen in Figure.\ref{fig:result}. It can be seen that the mouth area of the person appears blurry in frames generated by Wav2Lip. This is because regardless of the resolution of the input image, Wav2Lip always generates a face image of size 96$\times$96, which is then overlaid onto the original image. The overlay process can introduce artifacts or false details along the edges of the pasted area. The frames generated by Talklip exhibit similar issues to those of Wav2lip. MakeItTalk and SadTalk generate talking head videos by distorting a single image, resulting in the background being warped along with mouth movements, which compromises the authenticity of the generated videos. The frame generated by Difftalk exhibits high image quality. However, upon comparing the results in rows one and six of Figure.\ref{fig:result},  it's evident that certain frames exhibit mismatches between audio and lip movements. This discrepancy results in poor temporal coherence in the generated video. By contrast, our proposed model can generate natural talking head videos with precise lip-sync while ensuring high-quality frame generation and temporal coherence. Subsequent quantitative analysis will further support this assertion. For further video comparison results, please refer to the supplementary materials.

\noindent\textbf{Quantitative Analysis}.We conduct quantitative evaluations on several widely used metrics. The results are reported in Table.\ref{tab:cap1}. Benefiting from the strong performance of LDMs in generating high-quality images, our method achieves the best performance across all visual quality metrics. As shown in Figure.\ref{fig:result}, the quality of frames generated by our method far surpasses that of GAN-based models such as Wav2lip and Talklip. For lip synchronization metrics, both LSE-C and LSE-D are calculated using SyncNet \cite{b25}.    Since Wav2lip and Talklip utilize SyncNet as their discriminator during training, it is reasonable for these two methods to achieve the best performance in LSE-C and LSE-D. Compared to the rest of the methods, our method still maintains competitiveness in terms of lip synchronization metrics. In conclusion, our model outperforms existing state-of-the-art methods taking into account both quantitative and qualitative results.

\subsection{Ablation Study}
In the second stage of our model, the landmarks serve as a condition in the denoising process. The input for the landmark condition comprises two categories: the visual input and the coordinate input. In this section, we perform an ablation study on these two types of inputs, analyzing their impact on the temporal coherence of generated talking head videos.
There are three different settings: (1) abandoning the visual input of landmarks (\textbf{w/o visual}), (2) abandoning the coordinate input of landmarks (\textbf{w/o corr}), (3) our full model (\textbf{Full}). The results are shown in Table.\ref{tab:cap2}.

Without the visual input, which serves as the main guiding information in the denoising process, the model exhibits poor performance on both temporal metrics. The coordinate input of landmarks serves as 
auxiliary guiding information to optimize the lip shapes, ensuring better temporal coherence. After removing the coordinate input, the temporal metrics of the model show a slightly inferior performance compared to the full model. Different from our model that utilizes landmarks as the intermediate representation, Difftalk directly employs the given speech as guiding information in the denoising process. Our model significantly outperforms Difftalk in terms of temporal metrics. We introduce facial landmarks as the intermediate representation with the aim of having them serve as more robust guiding information than speech to mitigate the diversity in generated frames, thereby optimizing the temporal consistency in talking head videos. The ablation study demonstrates the rationality of our proposed model. The supplementary materials will offer additional video results for visualization purposes.

\section{CONCLUSION}
In this paper, We introduce a novel two-stage diffusion-based model that utilizes facial landmarks as an intermediate representation for high-quality, well-synchronized, and temporally coherent talking head generation. The landmarks generated in the first stage will serve as a condition in the second stage. By introducing landmarks as robust guiding information in the denoising process, our method significantly alleviates the jitter issues in diffusion-based models. Extensive experiments demonstrate the rationality and effectiveness of our proposed model.

\begin{table}[]
\centering
\caption{Quantitive results of the ablations study on temporal metrics. In addition to the ablation setting, we also compared our model with the diffusion-based model Difftalk in these two temporal metrics. We scaled up the results of tLP by a factor of one hundred and Pixel-MSE by a factor of one thousand.}
\label{tab:cap2}
\begin{tabular}{c|cc}
\hline
Method     & tLP$\downarrow$ $\times$ 100   & Pixel-MSE$\downarrow$ $\times$ 1000\\ \hline
DiffTalk   & 0.869 & 0.818    \\
w/o visual & 0.927 &  0.913    \\ 
w/o corr   & 0.553 & 0.751     \\ \hline
Full       & \textbf{0.495} &  \textbf{0.730}   \\ \hline
\end{tabular}
\end{table}

\section*{Acknowledgements}
This work was supported by the National Natural Science Foundation of China (62076100), Fundamental Research Funds for the Central Universities, SCUT (x2rjD2230080), the Science and Technology Planning Project of Guangdong Province (2020B0101100002), Guangdong Provincial Fund for Basic and Applied Basic Research - Regional Joint Fund Project (Key Project) (2023B1515120078), CAAI-Huawei MindSpore Open Fund, CCF-Zhipu AI Large Model Fund, Guangdong Provincial Natural Science Foundation for Outstanding Youth Team Project No.2024B1515040010.

\end{document}